\documentclass[11pt]{IEEEtran}
\usepackage[table,xcdraw]{xcolor} 

\usepackage{pgf,tikz}
\usepackage{mathrsfs}
\usepackage[linesnumbered,ruled,vlined]{algorithm2e}
\usepackage{tabularx} 
\usepackage{caption} 
\usepackage{pgfplots}
\pgfplotsset{compat=1.14}
\usetikzlibrary{arrows}
\usepackage{graphicx}
\usepackage{multirow}
\usepackage{subcaption} 
\providecommand{\keywords}[1]{\textbf{\textit{Keywords---}} #1}

\title{Data Augmentation by AutoEncoders for Unsupervised Anomaly Detection}
\author{Kasra Babaei, ZhiYuan Chen, Tomas Maul}

\begin{document}
\maketitle

\begin{abstract}This paper proposes an autoencoder (AE) that is used for improving the performance of once-class classifiers for the purpose of detecting anomalies. Traditional one-class classifiers (OCCs) perform poorly under certain conditions such as high-dimensionality and sparsity. Also, the size of the training set plays an important role on the performance of one-class classifiers. Autoencoders have been widely used for obtaining useful latent variables from high-dimensional datasets. In the proposed approach, the AE is capable of deriving meaningful features from high-dimensional datasets while doing data augmentation at the same time. The augmented data is used for training the OCC algorithms. The experimental results show that the proposed approach enhance the performance of OCC algorithms and also outperforms other well-known approaches.
\end{abstract}
\keywords{anomaly detection, autoencoder, data augmentation, one-class classifier}

\section{Introduction}
\par
\begin{figure}[!h]
\begin{center}
	
	\begin{tikzpicture}[scale=0.8,line cap=round,line join=round,>=triangle 45,x=1.0cm,y=1.0cm]
	\clip(-4.83199028842192,-3.350560119718393) rectangle (5.188918188289323,2.667106384609909);
	\draw [line width=1.2pt] (-4.,-2.) circle (0.32802438933713446cm);
	\draw [line width=1.2pt] (-4.,2.) circle (0.3124099870362663cm);
	\draw [line width=1.2pt] (-4.,1.) circle (0.32802438933713435cm);
	\draw [line width=1.2pt] (-4.,0.) circle (0.3124099870362663cm);
	\draw [line width=1.2pt] (-4.,-1.) circle (0.3124099870362663cm);
	\draw [line width=1.2pt] (-2.,1.) circle (0.3280243893371345cm);
	\draw [line width=1.2pt] (-2.,0.) circle (0.3124099870362662cm);
	\draw [line width=1.2pt] (-2.,-1.) circle (0.31240998703626616cm);
	\draw [line width=1.2pt] (0.,0.) circle (0.3124099870362662cm);
	\draw [line width=1.2pt] (2.,1.) circle (0.32802438933713435cm);
	\draw [line width=1.2pt] (2.,0.) circle (0.3124099870362663cm);
	\draw [line width=1.2pt] (2.,-1.) circle (0.3124099870362663cm);
	\draw [line width=1.2pt] (4.,-2.) circle (0.32802438933713446cm);
	\draw [line width=1.2pt] (4.,2.) circle (0.3124099870362663cm);
	\draw [line width=1.2pt] (4.,1.) circle (0.32802438933713435cm);
	\draw [line width=1.2pt] (4.,0.) circle (0.3124099870362663cm);
	\draw [line width=1.2pt] (4.,-1.) circle (0.3124099870362663cm);
	\draw [->,line width=0.8pt] (-3.671975610662866,1.) -- (-2.3272978670955706,0.9781801421936286);
	\draw [->,line width=0.8pt] (-3.6875900129637342,0.) -- (-2.312409987036266,0.);
	\draw [->,line width=0.8pt] (-3.687590012963734,-1.) -- (-2.311616059483331,-1.022258289963095);
	\draw [->,line width=0.8pt] (-1.6875900129637338,0.) -- (-0.3124099870362662,0.);
	\draw [->,line width=0.8pt] (0.3124099870362662,0.) -- (1.6727021329044296,0.9781801421936286);
	\draw [->,line width=0.8pt] (0.3124099870362662,0.) -- (1.6875900129637338,0.);
	\draw [->,line width=0.8pt] (0.3124099870362662,0.) -- (1.6903305043977526,-1.0412892660802997);
	\draw [->,line width=0.8pt] (2.3257776205028176,0.9616732211173156) -- (3.67270213290443,0.9781801421936286);
	\draw [->,line width=0.8pt] (2.312409987036266,0.) -- (3.6875900129637342,0.);
	\draw [->,line width=0.8pt] (2.312409987036266,0.) -- (3.672809219099011,-2.023370770064356);
	\draw [line width=0.8pt] (2.3118015931053515,-1.0194875995690844)-- (3.687590012963734,2.);
	\draw [->,line width=0.8pt] (2.3118015931053515,-1.0194875995690844) -- (3.687590012963734,-1.);
	\draw [line width=2.pt] (-4.580833035493351,-2.6108687372438015)-- (-0.9977224069976642,-2.6004223505718023);
	\draw [line width=2.pt] (0.976751104527751,-2.591669597353602)-- (4.559861733023438,-2.581223210681603);
	\draw [line width=2.pt] (-0.49585612643783666,-2.616490163162033)-- (0.5028204423655828,-2.616490163162033);
	\draw [line width=0.8pt] (-3.687590012963734,2.)-- (-2.3272978670955706,0.9781801421936286);
	\draw [line width=0.8pt] (-3.6875900129637342,0.)-- (-2.3272978670955706,0.9781801421936286);
	\draw [line width=0.8pt] (-3.687590012963734,-1.)-- (-2.3272978670955706,0.9781801421936286);
	\draw [line width=0.8pt] (-3.672614412416798,-2.02046159922395)-- (-2.3272978670955706,0.9781801421936286);
	\draw [line width=0.8pt] (-3.687590012963734,2.)-- (-2.312409987036266,0.);
	\draw [line width=0.8pt] (-3.671975610662866,1.)-- (-2.312409987036266,0.);
	\draw [line width=0.8pt] (-3.687590012963734,-1.)-- (-2.312409987036266,0.);
	\draw [line width=0.8pt] (-3.672614412416798,-2.02046159922395)-- (-2.312409987036266,0.);
	\draw [line width=0.8pt] (-3.687590012963734,2.)-- (-2.311616059483331,-1.022258289963095);
	\draw [line width=0.8pt] (-3.671975610662866,1.)-- (-2.311616059483331,-1.022258289963095);
	\draw [line width=0.8pt] (-3.672614412416798,-2.02046159922395)-- (-2.311616059483331,-1.022258289963095);
	\draw [line width=0.8pt] (-1.672480652362455,0.981804480686803)-- (-0.3124099870362662,0.);
	\draw [line width=0.8pt] (-1.688129116017763,-1.018345346116602)-- (-0.3124099870362662,0.);
	\draw [line width=0.8pt] (2.312409987036266,0.)-- (3.67270213290443,0.9781801421936286);
	\draw [line width=0.8pt] (2.3118015931053515,-1.0194875995690844)-- (3.67270213290443,0.9781801421936286);
	\draw [line width=0.8pt] (2.3257776205028176,0.9616732211173156)-- (3.6875900129637342,0.);
	\draw [line width=0.8pt] (2.3118015931053515,-1.0194875995690844)-- (3.6875900129637342,0.);
	\draw [line width=0.8pt] (2.3257776205028176,0.9616732211173156)-- (3.687590012963734,-1.);
	\draw [line width=0.8pt] (2.312409987036266,0.)-- (3.687590012963734,-1.);
	\draw [line width=0.8pt] (2.3257776205028176,0.9616732211173156)-- (3.672809219099011,-2.023370770064356);
	\draw [line width=0.8pt] (2.3118015931053515,-1.0194875995690844)-- (3.672809219099011,-2.023370770064356);
	\draw [line width=0.8pt] (2.3257776205028176,0.9616732211173156)-- (3.687590012963734,2.);
	\draw [->,line width=0.8pt] (2.319832405866583,-0.011039630830771835) -- (3.6847041472632425,1.993666296669693);
	\begin{scriptsize}
	\draw[color=black] (-2.7,-2.8) node {$Encoder$};
	\draw[color=black] (2.8,-2.8) node {$Decoder$};
	\draw[color=black] (0.0,-2.8) node {$Bottleneck$};
	\end{scriptsize}
	\end{tikzpicture}
\end{center}
\caption{The structure of a deep undercomplete autoencoder}
\label{fig:structure_of_ae}
\end{figure}
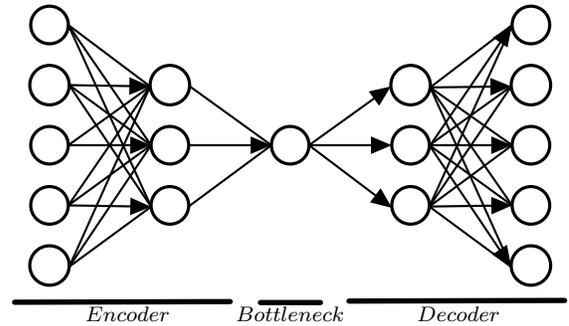

 Deep neural networks have demonstrated their effectiveness and managed to improve the state-of-the-art in diverse application areas. In particular, AutoEncoders (AEs), also previously known as auto-associative neural networks, have shown to be a very powerful tool for the purpose of reducing dimensionality thanks to their ability in discovering non-linear correlations between features \cite{8363930}. This important capability of AutoEncoders has made them very suitable for the task of anomaly detection. AutoEncoders come in various architectures. When the number of features in the middle layer, known as the bottleneck, is smaller than the number of inputs, as depicted in Fig. \ref{fig:structure_of_ae}, the high dimensional space can be transformed into a low dimensional space \cite{wang2016auto}. AutoEncoders are different from other dimensionality reduction methods such as Principal Component Analysis (PCA) in the sense that AEs generally perform non-linear dimensionality reduction. Moreover, according to the literature, AEs generally outperform PCA \cite{wang2016auto}. Besides, statistical approaches such as PCA or Zero Component Analysis (ZCA) require more memory to calculate the covariance matrix as the dimensionality increases \cite{yousefi2017autoencoder}.

\par AutoEncoders have been employed for the task of anomaly detection in various ways. As the network tries to reconstruct each data point, it generates a value, known as reconstruction error (RE), that shows the degree of resemblance between the original input and its reconstruction at the output layer. In some studies such as \cite{aygun2017network}, authors used the reconstruction error to create a threshold-based classifier for separating normal points from anomalies. It is also possible to use an AE to reduce the dimensionality and pass the latent variables obtained from the bottleneck to a One-Class Classification (OCC) algorithm such as One-Class SVM (OCSVM). The authors in \cite{Cao20193074} applied a number of OCC algorithms such as Local Outlier Factor (LOF) and Kernel Density Estimation (KDE) on the latent variables that were obtained from a regularised AE to capture anomalies. 

\par There are various challenges in the area of anomaly detection including availability of anomalous examples and imbalanced class distribution \cite{Chandola2009}, which debilitate performance. Many real-world datasets suffer from the class imbalance problem. In a skewed dataset, for instance in a binary dataset, each data point belongs to either the majority class or the minority class. The majority class contains a greater number of data points than the minority class \cite{OH20191}. The ratio between the majority class and the minority class is known as the imbalance ratio and varies from one domain to another \cite{Barua2014405}. 

\par The imbalanced dataset problem weakens the performance and can cause misclassification \cite{Chandola2009}. In particular, supervised approaches suffer more because it is hard to obtain examples of anomalies for training the model or the current definition of anomaly changes over time. Therefore, unsupervised models or OCC algorithms appear to be more suitable. In an OCC algorithm, the model is trained with a training set that only includes data instances from one class (known as the target class) and the model is expected to separate data instances of the target class from non-target class instances in the test set \cite{Khan2010}. In order to employ an OCC, it is necessary to have access to a training set that includes merely normal examples. It is worth noting that in some scenarios the training set includes a small number of data points from other classes as well. This training set is used to set the threshold at which non-target and target points will be divided. In an anomaly detection problem, the goal is to separate anomalies from normal points.

\par There are several methods to overcome the issue of imbalanced class distributions that can be categorised into data-level, algorithmic-level, and cost-sensitive methods \cite{Douzas2018}. In a data-level method, the goal is to bring balance to the dataset prior to performing classification by oversampling, undersampling or a hybrid approach \cite{Domingos:1999}. Two widely used oversampling methods are Synthetic Minority Oversampling Technique (SMOTE) and adaptive synthetic sampling (ADASYN) \cite{HeH.BaiY.GarciaE.&Li2008}. Authors in \cite{OH20191} proposed a new approach in which a Generative Adversarial Network (GAN) is used to enhance classification performance in imbalanced datasets. They used a variant of GAN that can also detect outliers in the majority class, which prevents creating a biased classification boundary.  In a similar approach, authors in \cite{Douzas2018} proposed a conditional Generative Adversarial Network (cGAN) for oversampling the minority class in a binary classification problem. In their approach, the network is conditioned on external information, i.e., class labels, to approximate the actual class distribution.

\par The performance of OCC methods depends on various factors including the size of the training set. These methods can perform better when the training set includes more data points as it makes it feasible to compute a less ambiguous class boundary for dividing data points of the target class from the rest \cite{Khan2010}. Data augmentation refers to the process in which the training instances are oversampled to improve the model's performance \cite{ghaffar2019data}. This approach is widely used in machine learning tasks. In the area of anomaly detection, the authors in \cite{8594955} claim to be the first to use data augmentation in an unsupervised approach for detecting anomalies. They employed a variant of a GAN to obtain latent variables, and selectively oversampled instances close to the head and tail of the latent variables by an approach similar to SMOTE. They argued that their approach addresses the problem of scarcity of infrequent normal instances which can reduce the performance of density-based anomaly detection algorithms.

\par This paper is motivated by the question of whether AEs are capable of augmenting data points with meaningful features for the purpose of unsupervised anomaly detection or not. To explain further, the latent variables of the bottleneck of the AE are collected from a certain number of epochs and then used to train various OCC algorithms for finding anomalies. An extensive amount of experimentation was carried out to demonstrate that this approach can lead to achieving better anomaly detection performance from OCC algorithms. 

\section{AutoEncoders}
\label{section:AutoEncoders}
\par In this section, a summary of the AutoEncoder (AE) neural network is presented. The AE is an unsupervised multi-layer neural network with two fundamental components, namely the \textit{encoder} and the \textit{decoder}. While the encoder tries to transform high dimensional data into a lower dimension, the decoder attempts to approximately reconstruct the input from the low-dimensional feature space. The difference between the input and the reconstructed data point is known as the reconstruction error and by training the network, the AE tries to minimise this error, i.e., to maximise the resemblance between the reconstructed data and the original input. 

\begin{figure}[!h]
	\begin{center}
		\begin{tikzpicture}[line cap=round,line join=round,>=triangle 45,x=1.0cm,y=1.0cm]
		\clip(-3,-0.6) rectangle (3,.6);
		\draw [line width=0.8pt] (-2.,0.) circle (0.5);
		\draw [line width=0.8pt] (0.,0.) circle (0.5);
		\draw [line width=0.8pt] (2.,0.) circle (0.5);
		\draw [->,line width=1.pt] (-1.5,0.) -- (-0.5,0.);
		\draw [->,line width=1.pt] (0.5,0.) -- (1.5,0.);
		\begin{scriptsize}
		\draw[color=black] (-2,0) node {$x$};
		\draw[color=black] (0,0) node {$z$};
		\draw[color=black] (2,0) node {$y$};
		
		\draw[color=black] (-1.2,0.2) node {$u$};
		\draw[color=black] (0.85,0.2) node {$v$};
		\end{scriptsize}
		\end{tikzpicture}
	\end{center}
	\caption{The structure of a basic autoencoder}
	\label{fig:basic_ae}
\end{figure}
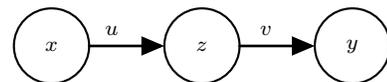

\par As shown in Figure \ref{fig:basic_ae}, the most generic AE with only one latent layer attempts to transform the input $x$ into latent vector $z$ using an encoder represented by function $u$. Then, the decoder tries to map $z$ to reconstruct $y$ by using a decoder represented as function $v$. Having a training set, $D_u=\{x_1,x_2, x_3,...,x_n\}$, where $n$ refers to the number of instances in $D$ and $x_i$ is the $i$th instance with $m$ features, the encoder can then be defined as:
\begin{equation}
	z = u(x) = s(W x + b)
	\label{equ:encoder}
\end{equation}
while the decoder can be defined as:
\begin{equation}
	y = v(z) = s'(W' z + b')
	\label{equ:decoder}
\end{equation}
where both $s$ and $s'$ represent the activation functions, $W$ and $W'$ denote the weight matrices while $b$ and $b'$ represent the bias vectors. Choosing the right activation function depends on various factors; however, often a non-linear activation function such as Sigmoid or ReLU can capture more useful representations.  

\par The AE has several variants. In an under-complete AE where the number of nodes in the middle layer is smaller than the number of nodes at the input layer, the aim is to reduce the input dimension by learning some desired latent features. A detailed review of different variants of AEs can be found in \cite{charte2018practical}. 

\section{SMOTE and ADASYN}
\label{sec:smote_adasyn}
\par This section briefly explains the two oversampling approaches used against the proposed method in this paper for comparison purposes. 

\par In Synthetic Minority Oversampling Technique (SMOTE), the skewed dataset is rebalanced simply by replicating the minority class. This is done by generating synthetic data points on lines between the data points within a defined neighbourhood; therefore, this approach concentrates on the feature space rather than on the data space \cite{Fernandez2018}. 

\par In the Adaptive Synthetic (ADASYN) sampling approach, which is an extension of SMOTE \cite{OH20191}, the difficulty of learning a data point is taken into account and data points with a higher level of difficulty in the minority class are replicated more to rebalance the data set \cite{HeH.BaiY.GarciaE.&Li2008}. This approach employs a weighted distribution for each data point in the minority class which conforms to the learning difficulty of the respective example.

\section{Proposed Method}
\label{section:proposed_method}
\par In this section, the proposed method is elaborated upon. The purpose of this approach is to use an AE to augment the training set for the purpose of detecting anomalies. This augmented training set is later used in an OCC method that tries to detect anomalies in the test set. The performance of the OCC depends on how well its threshold is defined. The assumption here is that by augmenting the training set, it is possible to compute a better class boundary and ultimately improve the performance of the OCC. 

\par Having a training set, the AE tries to minimise the reconstruction error by updating its weights through error backpropagation. Traditionally, once the network is fully trained, latent variables are extracted from the bottleneck of the AE and used to train the OCC algorithm. However, in the proposed approach, the latent variables of the last $\nu$ epochs are extracted and concatenated to form an augmented training set. The value of $\nu$ is a parameter that needs to be tuned. The assumption is that the network is almost well trained during the last few epochs and that the corresponding latent variables possess a good representation of the original input. The algorithm of the proposed approach is summarised in Algorithm \ref{alg:AEOS}, in which $n' > n$ as the model is augmenting the training set while $m' < m$ as at the dimensionality is being reduced. Another advantage of the proposed model is that unlike other approaches, it is not necessary to perform extra computation for augmenting the training set.


\begin{algorithm}[h]

\SetAlgoLined
 
 \SetKwInOut{Input}{input}\SetKwInOut{Output}{output}
 \Input{$D$: a dataset with $n\times m$ dimension}

 \Output{$D'$: a dataset with $n'\times m'$ dimension} 
 \BlankLine

$n\_epochs \leftarrow$ number of epochs\;
$\nu \leftarrow$ portion of epochs to use\;

 \BlankLine
  
\For{each epoch $i$ in $n\_epochs$}{
	Reconstruct the input\;
	Compute the reconstruction error\;
	Backpropagate the loss\;
	Take a gradient step\;
	\If {$i \geq ((1-\nu)\times n\_epochs)$} { 
		$z \leftarrow$ latent variables in the bottleneck\;
		$D' \leftarrow$ append $z$\;
	}	
 }
\Return $D'$\; 	
\caption{Over-sampling by AE}
\label{alg:AEOS}
\end{algorithm}

\section{Evaluation and Discussion}
\label{section:evaluation_and_discussion}
\par This section presents the evaluation of the proposed approach for augmenting the training set in order to improve the performance of anomaly detection with OCC algorithms. To demonstrate the effectiveness of the proposed approach, three OCC algorithms, namely Local Outlier Factor (LOF), Kernel Density Estimation (KDE), and Isolation Forest (ISF) were used in the experiments. 

\subsection{Datasets}
\label{subsec:datasets}

\begin{table}[]
\caption{Details of the 8 data sets used in the experiments}
\label{table:data_sets_details}
\resizebox{\columnwidth}{!}{%
\begin{tabular}{l|c|c|c|c}
Data set name & \begin{tabular}[c]{@{}c@{}}No. of \\ features\end{tabular} & \begin{tabular}[c]{@{}c@{}}Training \\ set\end{tabular} & \begin{tabular}[c]{@{}c@{}}Validation \\ set\end{tabular} & \begin{tabular}[c]{@{}c@{}}Testing \\ set\end{tabular} \\ \hline
PenDigits & 16 & 624 & 312 & 727 \\
Shuttle & 9 & 27286 & 8701 & 14500 \\
Spambase & 57 & 2230 & 459 & 460 \\
InternetAds & 1558 & 1582 & 235 & 236 \\
Arrhythmia & 259 & 142 & 83 & 84 \\
\hline
NSLKDD & 122 & 5356 & 1571 & 12132 \\
UNSW & 196 & 4488 & 1478 & 43062 \\
Virut(CTU13-13) & 40 & 19163 & 14387 & 14389 \\
Rbot(CTU13-10) & 38 & 9508 & 24439 & 24439 \\
Neris(CTU13-09) & 41 & 17980 & 42990 & 42990 \\
Murlo(CTU13-08) & 40 & 43693 & 15789 & 15789
\end{tabular}%
}
\end{table}

\par The experiments were carried out on 8 publicly available data sets which are widely used in this domain. The details of the data sets are shown in Table \ref{table:data_sets_details}. As depicted in the table, the first five data sets possess a single-type of anomaly while the rest (NSL-KDD, UNSW-NB15 and CTU13) have different types of anomalies. The single-type anomaly data sets, i.e., PenDigits, Shuttle, Spambase, and InternetAds, are from the UCI machine learning repository \cite{UCI_repo}. 

\par The CUT13 dataset is a well-known botnet data set \cite{GARCIA2014100}. In this experiment, four out of 13 botnet scenarios were selected in which 60\% of the normal traffic was used for training, 20\% for validation, and the last 20\% of the normal and botnet traffic were used for testing. This data set includes three categorical features that are encoded by one-hot-encoding, which ultimately increased the number of features to what is listed in Table \ref{table:data_sets_details}. 

\par The NSL-KDD data set is a modified version of the KDD'99 data set which has resolved some of the intrinsic problems with the original version \cite{Cao20193074}. One of the major issues with the original dataset is a large number of redundant records, which have been removed in NSL-KDD \cite{tavallaee2009}. The anomalies in this data set belong to one of the following four attacks: 1) Denial of Service (DoS); 2) Remote to Local (R2L); 3) User to Local (U2R); and 4) Probe. Similar to CTU, this data set also includes three categorical features that are encoded by one-hot-encoding, which increased the number of dimensions from 14 to 122. 

\par The UNSW-NB15 data set is a network data set with nine types of attack: 1) fuzzers; 2) analysis; 3) backdoor; 4) DoS; 5) exploit; 6) generic; 7) reconnaissance; 8) shell-code; and 9) worm \cite{UNSW-NB15}. The categorical features are encoded by the same method which increased the number of features from 47 to 196. 

\par It is worth mentioning that as suggested by Van at al. \cite{Cao20193074}, the training sets of UNSW-NB15 and NSL-KDD are considerably larger than other data sets; therefore, only 10\% of the training sets were used (as tabulated in Table \ref{table:data_sets_details}). All the data sets were normalised to the range $[-1,1]$ and correspondingly the tangent activation function was used in the network. The parameter $\nu$ was experimentally tuned to $0.25$. 

\subsection{AE Parameters}
\label{subsection:ae_parameters}
\par The loss function used in this experiment was $Smoothl1Loss$, which is essentially a combination of $L2$ and $L1$ terms, i.e., it uses $L2$ if the absolute element-wise error is less than 1 and $L1$ term if not. To minimise the loss function, Stochastic Gradient Descent (SGD) was used except for six data sets, i.e., Arrhythmia, Murlo, Rbot, InternetAds, Shuttle, and Spambase in which the Root Mean Square Propagation (RMSProp) method was selected. This decision was made due to the fact that by changing the optimiser it became possible to avoid overfitting in every approach. The number of layers was set to 5 for all the data sets as suggested in \cite{ERFANI2016121} with the number of nodes in the bottleneck being set to $m=[1+\sqrt{n}]$, where $n$ is the number of features of the input \cite{CaoVan2016}. For data sets with more than $2000$ instances, the mini-batch size was set to $64$ and $16$ otherwise. To avoid overfitting, a simple early stopping heuristic was implemented to stop the training process when the network was no longer learning after a certain number of iterations or when the learning improvement was insignificant. 

\subsection{OCC Parameters}
\label{subsection:occ_parameters}
\par In Isolation Forest, the number of estimators was set to 20. In LOF, the number of neighbours was set to 20 while the contamination rate, i.e., the ratio between anomalies and normal examples, was set to $0.1$. This contamination rate and the training set was used to compute a class boundary during the training phase.

\begin{table*}[h]
\centering
\caption{PR AUC and ROC AUC for the UNSW data set}
\label{tab:aucs_UNSW}
\resizebox{\textwidth}{!}{%
\begin{tabular}{l|l|c|c|c|c|c|c|c|c|c|c|c|c|c|c|c|c|cc}
\multirow{2}{*}{\begin{tabular}[c]{@{}l@{}}Data augmentation \\ method\end{tabular}} & \multirow{2}{*}{\begin{tabular}[c]{@{}l@{}}OCC \\ method\end{tabular}} & \multicolumn{2}{c|}{Fuzzers} & \multicolumn{2}{c|}{Analysis} & \multicolumn{2}{c|}{Backdoor} & \multicolumn{2}{c|}{DoS} & \multicolumn{2}{c|}{Exploits} & \multicolumn{2}{c|}{Generic} & \multicolumn{2}{c|}{Reconnaissance} & \multicolumn{2}{c|}{Shellcode} & \multicolumn{2}{c}{Worms} \\ \cline{3-20} 
 &  & PR & ROC & PR & ROC & PR & ROC & PR & ROC & PR & ROC & PR & ROC & PR & ROC & PR & ROC & \multicolumn{1}{c|}{PR} & ROC \\ \hline\hline
\multirow{3}{*}{None} & ISF & 0.825 & 0.478 & 0.995 & 0.782 & 0.996 & 0.813 & 0.949 & 0.689 & 0.758 & 0.502 & 0.939 & 0.878 & 0.914 & 0.531 & 0.990 & 0.476 & \multicolumn{1}{c|}{\cellcolor[HTML]{9B9B9B}0.999} & 0.603 \\ \cline{2-20} 
 & KDE & 0.821 & 0.456 & 0.993 & 0.800 & 0.995 & 0.787 & 0.956 & 0.682 & 0.744 & 0.412 & 0.810 & 0.671 & 0.921 & 0.552 & 0.992 & 0.519 & \multicolumn{1}{c|}{0.998} & 0.421 \\ \cline{2-20} 
 & LOF & 0.837 & 0.437 & \cellcolor[HTML]{9B9B9B}0.996 & 0.847 & 0.985 & 0.475 & 0.843 & 0.402 & 0.693 & 0.515 & 0.921 & 0.552 & 0.877 & 0.349 & 0.985 & 0.332 & \multicolumn{1}{c|}{0.998} & 0.328 \\ \hline\hline
\multirow{3}{*}{SMOTE} & ISF & 0.811 & 0.390 & 0.992 & 0.718 & 0.995 & 0.844 & 0.942 & 0.664 & 0.838 & 0.624 & 0.929 & 0.834 & 0.909 & 0.467 & 0.984 & 0.280 & \multicolumn{1}{c|}{\cellcolor[HTML]{9B9B9B}0.999} & \cellcolor[HTML]{9B9B9B}0.748 \\ \cline{2-20} 
 & KDE & 0.794 & 0.393 & 0.989 & 0.685 & 0.993 & 0.726 & 0.945 & 0.718 & 0.785 & 0.527 & 0.927 & 0.817 & 0.928 & 0.502 & 0.987 & 0.349 & \multicolumn{1}{c|}{0.997} & 0.363 \\ \cline{2-20} 
 & LOF & 0.792 & 0.285 & 0.962 & 0.240 & 0.971 & 0.311 & 0.853 & 0.436 & 0.710 & 0.466 & 0.718 & 0.622 & 0.841 & 0.275 & 0.982 & 0.295 & \multicolumn{1}{c|}{0.996} & 0.104 \\ \hline\hline
\multirow{3}{*}{ADASYN} & ISF & 0.848 & 0.489 & 0.992 & 0.810 & 0.995 & 0.784 & 0.964 & 0.748 & 0.718 & 0.427 & 0.921 & 0.822 & 0.894 & 0.525 & 0.985 & 0.470 & \multicolumn{1}{c|}{\cellcolor[HTML]{9B9B9B}0.999} & 0.475 \\ \cline{2-20} 
 & KDE & 0.848 & 0.489 & 0.992 & 0.810 & 0.995 & 0.784 & 0.964 & 0.748 & 0.718 & 0.427 & 0.921 & 0.822 & 0.894 & 0.525 & 0.985 & 0.470 & \multicolumn{1}{c|}{\cellcolor[HTML]{9B9B9B}0.999} & 0.475 \\ \cline{2-20} 
 & LOF & 0.837 & 0.440 & 0.969 & 0.376 & 0.987 & 0.585 & 0.834 & 0.352 & 0.734 & 0.385 & 0.662 & 0.523 & 0.891 & 0.346 & 0.986 & 0.341 & \multicolumn{1}{c|}{0.998} & 0.527 \\ \hline\hline
\multirow{3}{*}{Noise} & ISF & 0.828 & 0.480 & 0.994 & 0.792 & 0.994 & 0.737 & 0.965 & 0.776 & 0.864 & \cellcolor[HTML]{9B9B9B}0.693 & 0.929 & 0.856 & 0.938 & 0.555 & 0.994 & 0.620 & \multicolumn{1}{c|}{0.998} & 0.366 \\ \cline{2-20} 
 & KDE & 0.896 & 0.506 & 0.992 & 0.817 & 0.996 & 0.833 & 0.917 & 0.627 & 0.798 & 0.612 & 0.933 & 0.874 & 0.944 & 0.562 & 0.989 & 0.473 & \multicolumn{1}{c|}{\cellcolor[HTML]{9B9B9B}0.999} & 0.632 \\ \cline{2-20} 
 & LOF & 0.850 & 0.483 & 0.990 & 0.622 & 0.987 & 0.539 & 0.864 & 0.383 & 0.731 & 0.462 & 0.842 & 0.784 & 0.949 & \cellcolor[HTML]{9B9B9B}0.721 & 0.991 & 0.488 & \multicolumn{1}{c|}{\cellcolor[HTML]{9B9B9B}0.999} & 0.484 \\ \hline\hline
\multirow{3}{*}{Proposed method} & ISF & 0.892 & 0.566 & \cellcolor[HTML]{9B9B9B}0.996 & \cellcolor[HTML]{9B9B9B}0.851 & \cellcolor[HTML]{9B9B9B}0.998 & \cellcolor[HTML]{9B9B9B}0.924 & 0.970 & 0.826 & \cellcolor[HTML]{9B9B9B}0.873 & 0.681 & 0.970 & 0.943 & 0.948 & 0.591 & \cellcolor[HTML]{9B9B9B}0.995 & \cellcolor[HTML]{9B9B9B}0.680 & \multicolumn{1}{c|}{\cellcolor[HTML]{9B9B9B}0.999} & 0.713 \\ \cline{2-20} 
 & KDE & \cellcolor[HTML]{9B9B9B}0.932 & \cellcolor[HTML]{9B9B9B}0.668 & 0.995 & 0.814 & \cellcolor[HTML]{9B9B9B}0.998 & 0.905 & \cellcolor[HTML]{9B9B9B}0.974 & 0.821 & 0.868 & 0.649 & \cellcolor[HTML]{9B9B9B}0.975 & \cellcolor[HTML]{9B9B9B}0.959 & \cellcolor[HTML]{9B9B9B}0.954 & 0.641 & 0.988 & 0.467 & \multicolumn{1}{c|}{\cellcolor[HTML]{9B9B9B}0.999} & 0.605 \\ \cline{2-20} 
 & LOF & 0.875 & 0.540 & 0.992 & 0.801 & 0.995 & 0.842 & 0.972 & \cellcolor[HTML]{9B9B9B}0.830 & 0.792 & 0.642 & 0.861 & 0.787 & \cellcolor[HTML]{9B9B9B}0.954 & 0.718 & 0.990 & 0.496 & \multicolumn{1}{c|}{0.995} & 0.133 
\end{tabular}%
}
\end{table*}
\begin{table*}[h]
\centering
\caption{PR AUC and ROC AUC for the NSL-KDD and CTU13 data set}
\label{tab:aucs_nlkdd_ctu13}
\resizebox{\textwidth}{!}{%
\begin{tabular}{l|l|c|c|c|c|c|c|c|c|c|c|c|c|c|c|c|cc}
\multirow{2}{*}{\begin{tabular}[c]{@{}l@{}}Data augmentation \\ method\end{tabular}} & \multirow{2}{*}{\begin{tabular}[c]{@{}l@{}}OCC \\ method\end{tabular}} & \multicolumn{2}{c|}{Probe} & \multicolumn{2}{c|}{DoS} & \multicolumn{2}{c|}{R2L} & \multicolumn{2}{c|}{U2R} & \multicolumn{2}{c|}{Virut} & \multicolumn{2}{c|}{Rbot} & \multicolumn{2}{c|}{Neris} & \multicolumn{2}{c}{Murlo} \\ \cline{3-18} 
 &  & PR & ROC & PR & ROC & PR & ROC & PR & ROC & PR & ROC & PR & ROC & PR & ROC & \multicolumn{1}{c|}{PR} & ROC \\ \hline\hline
\multirow{3}{*}{No oversampling} & ISF & 0.982 & 0.927 & 0.830 & 0.808 & 0.839 & 0.691 & \cellcolor[HTML]{9B9B9B}0.999 & 0.828 & 0.384 & 0.100 & 0.783 & 0.024 & 0.749 & 0.100 & \multicolumn{1}{c|}{0.064} & 0.094 \\ \cline{2-18} 
 & KDE & 0.975 & 0.893 & 0.828 & 0.801 & 0.824 & 0.667 & \cellcolor[HTML]{9B9B9B}0.999 & 0.821 & 0.364 & 0.092 & 0.704 & 0.046 & 0.746 & 0.270 & \multicolumn{1}{c|}{0.041} & 0.068 \\ \cline{2-18} 
 & LOF & 0.683 & 0.188 & 0.539 & 0.535 & 0.657 & 0.179 & 0.981 & 0.140 & 0.840 & 0.809 & 0.718 & 0.082 & 0.882 & 0.426 & \multicolumn{1}{c|}{0.108} & 0.671 \\ \hline\hline
\multirow{3}{*}{SMOTE} & ISF & 0.976 & 0.900 & 0.728 & 0.765 & 0.950 & 0.829 & 0.997 & 0.715 & 0.374 & 0.067 & 0.776 & 0.030 & 0.738 & 0.155 & \multicolumn{1}{c|}{0.696} & 0.797 \\ \cline{2-18} 
 & KDE & 0.967 & 0.880 & 0.795 & 0.863 & 0.938 & 0.785 & 0.997 & 0.698 & 0.370 & 0.116 & 0.698 & 0.018 & 0.756 & 0.260 & \multicolumn{1}{c|}{0.061} & 0.297 \\ \cline{2-18} 
 & LOF & 0.627 & 0.082 & 0.382 & 0.132 & 0.626 & 0.187 & 0.980 & 0.078 & 0.585 & 0.651 & 0.718 & 0.065 & 0.834 & 0.312 & \multicolumn{1}{c|}{0.233} & 0.865 \\ \hline\hline
\multirow{3}{*}{ADASYN} & ISF & 0.831 & 0.574 & 0.872 & 0.924 & 0.906 & 0.736 & 0.998 & 0.762 & 0.374 & 0.067 & 0.729 & 0.017 & 0.729 & 0.117 & \multicolumn{1}{c|}{0.218} & 0.225 \\ \cline{2-18} 
 & KDE & 0.953 & 0.862 & 0.856 & 0.910 & 0.938 & 0.775 & 0.997 & 0.808 & 0.364 & 0.089 & 0.696 & 0.004 & 0.751 & 0.286 & \multicolumn{1}{c|}{0.117} & 0.682 \\ \cline{2-18} 
 & LOF & 0.625 & 0.097 & 0.369 & 0.070 & 0.636 & 0.222 & 0.977 & 0.047 & 0.455 & 0.312 & 0.696 & 0.004 & 0.750 & 0.222 & \multicolumn{1}{c|}{0.044} & 0.166 \\ \hline\hline
\multirow{3}{*}{Noise} & ISF & 0.940 & 0.806 & 0.897 & 0.858 & 0.926 & 0.785 & 0.991 & 0.468 & 0.504 & 0.261 & 0.859 & 0.039 & 0.824 & 0.344 & \multicolumn{1}{c|}{0.213} & 0.776 \\ \cline{2-18} 
 & KDE & 0.965 & 0.885 & 0.857 & 0.833 & 0.948 & 0.816 & 0.997 & 0.737 & 0.433 & 0.287 & 0.701 & 0.029 & 0.903 & 0.509 & \multicolumn{1}{c|}{0.047} & 0.143 \\ \cline{2-18} 
 & LOF & 0.670 & 0.165 & 0.434 & 0.256 & 0.629 & 0.178 & 0.981 & 0.084 & 0.898 & \cellcolor[HTML]{9B9B9B}0.873 & 0.838 & 0.612 & 0.890 & 0.475 & \multicolumn{1}{c|}{0.547} & 0.878 \\ \hline\hline
\multirow{3}{*}{Proposed method} & ISF & 0.986 & 0.941 & 0.930 & 0.917 & 0.977 & \cellcolor[HTML]{9B9B9B}0.918 & \cellcolor[HTML]{9B9B9B}0.999 & \cellcolor[HTML]{9B9B9B}0.881 & 0.619 & 0.535 & 0.946 & 0.668 & 0.834 & 0.455 & \multicolumn{1}{c|}{0.504} & 0.904 \\ \cline{2-18} 
 & KDE & \cellcolor[HTML]{9B9B9B}0.991 & \cellcolor[HTML]{9B9B9B}0.972 & \cellcolor[HTML]{9B9B9B}0.954 & \cellcolor[HTML]{9B9B9B}0.931 & \cellcolor[HTML]{9B9B9B}0.969 & 0.887 & 0.998 & 0.862 & 0.491 & 0.391 & 0.834 & 0.617 & 0.928 & 0.600 & \multicolumn{1}{c|}{\cellcolor[HTML]{9B9B9B}0.763} & 0.797 \\ \cline{2-18} 
 & LOF & 0.745 & 0.494 & 0.570 & 0.590 & 0.798 & 0.593 & 0.987 & 0.423 & \cellcolor[HTML]{9B9B9B}0.906 & \cellcolor[HTML]{9B9B9B}0.873 & \cellcolor[HTML]{9B9B9B}0.993 & \cellcolor[HTML]{9B9B9B}0.961 & \cellcolor[HTML]{9B9B9B}0.928 & \cellcolor[HTML]{9B9B9B}0.682 & \multicolumn{1}{c|}{0.526} & \cellcolor[HTML]{9B9B9B}0.922
\end{tabular}%
}
\end{table*}
\begin{table*}[h]
\centering
\caption{PR AUC and ROC AUC for data sets with single-type anomaly }
\label{tab:aucs_single_type}
\resizebox{\textwidth}{!}{%
\begin{tabular}{l|l|c|c|c|c|c|c|c|c|cc}
\multirow{2}{*}{\begin{tabular}[c]{@{}l@{}}Data augmentation \\ method\end{tabular}} & \multirow{2}{*}{\begin{tabular}[c]{@{}l@{}}OCC \\ method\end{tabular}} & \multicolumn{2}{c|}{PenDigits} & \multicolumn{2}{c|}{Shuttle} & \multicolumn{2}{c|}{Spambase} & \multicolumn{2}{c|}{InternetAds} & \multicolumn{2}{c}{Arrhythmia} \\ \cline{3-12} 
 &  & PR & ROC & PR & ROC & PR & ROC & PR & ROC & \multicolumn{1}{c|}{PR} & ROC \\ \hline\hline
\multirow{3}{*}{No oversampling} & ISF & 0.694 & 0.712 & 0.714 & 0.440 & 0.296 & 0.327 & 0.160 & 0.525 & \multicolumn{1}{c|}{0.440} & 0.500 \\ \cline{2-12} 
 & KDE & 0.878 & 0.826 & 0.821 & 0.712 & 0.317 & 0.398 & 0.182 & 0.527 & \multicolumn{1}{c|}{0.416} & 0.446 \\ \cline{2-12} 
 & LOF & 0.646 & 0.777 & 0.759 & 0.427 & 0.293 & 0.311 & 0.221 & 0.604 & \multicolumn{1}{c|}{0.416} & 0.446 \\ \hline\hline
\multirow{3}{*}{SMOTE} & ISF & 0.557 & 0.371 & 0.784 & 0.511 & 0.341 & 0.461 & 0.134 & 0.411 & \multicolumn{1}{c|}{0.438} & 0.443 \\ \cline{2-12} 
 & KDE & 0.602 & 0.587 & 0.628 & 0.130 & 0.318 & 0.351 & 0.171 & 0.525 & \multicolumn{1}{c|}{0.440} & 0.50 \\ \cline{2-12} 
 & LOF & 0.319 & 0.103 & 0.711 & 0.364 & 0.314 & 0.332 & 0.165 & 0.489 & \multicolumn{1}{c|}{0.419} & 0.461 \\ \hline\hline
\multirow{3}{*}{ADASYN} & ISF & 0.725 & 0.715 & 0.727 & 0.400 & 0.313 & 0.376 & 0.168 & 0.525 & \multicolumn{1}{c|}{0.373} & 0.365 \\ \cline{2-12} 
 & KDE & 0.454 & 0.452 & 0.690 & 0.356 & 0.348 & 0.400 & 0.148 & 0.443 & \multicolumn{1}{c|}{0.377} & 0.392 \\ \cline{2-12} 
 & LOF & 0.388 & 0.215 & 0.763 & 0.597 & 0.331 & 0.386 & 0.127 & 0.374 & \multicolumn{1}{c|}{0.376} & 0.382 \\ \hline\hline
\multirow{3}{*}{Noise} & ISF & 0.916 & 0.867 & 0.923 & 0.792 & 0.491 & 0.622 & 0.221 & 0.660 & \multicolumn{1}{c|}{0.468} & 0.543 \\ \cline{2-12} 
 & KDE & 0.893 & 0.535 & 0.877 & \cellcolor[HTML]{9B9B9B}0.808 & 0.400 & 0.547 & 0.168 & 0.567 & \multicolumn{1}{c|}{0.460} & 0.604 \\ \cline{2-12} 
 & LOF & 0.491 & 0.415 & 0.846 & 0.718 & 0.464 & 0.629 & 0.141 & 0.451 & \multicolumn{1}{c|}{0.534} & 0.554 \\ \hline\hline
\multirow{3}{*}{Proposed method} & ISF & 0.947 & 0.918 & \cellcolor[HTML]{9B9B9B}0.926 & 0.781 & 0.683 & 0.821 & \cellcolor[HTML]{9B9B9B}0.428 & \cellcolor[HTML]{9B9B9B}0.738 & \multicolumn{1}{c|}{0.573} & 0.591 \\ \cline{2-12} 
 & KDE & \cellcolor[HTML]{9B9B9B}0.960 & \cellcolor[HTML]{9B9B9B}0.932 & 0.872 & 0.723 & \cellcolor[HTML]{9B9B9B}0.714 & \cellcolor[HTML]{9B9B9B}0.832 & 0.255 & 0.694 & \multicolumn{1}{c|}{0.583} & 0.604 \\ \cline{2-12} 
 & LOF & 0.930 & 0.908 & 0.890 & 0.744 & 0.614 & 0.761 & 0.368 & 0.719 & \multicolumn{1}{c|}{\cellcolor[HTML]{9B9B9B}0.604} & \cellcolor[HTML]{9B9B9B}0.616
\end{tabular}%
}
\end{table*} 

\subsection{Analysis and Discussion}
\label{subsection:evaluation_metrics}
\par The performance of the proposed approach is compared to the state-of-the-art. In particular, the two widely used oversampling methods, namely SMOTE and ADASYN, were used to increase the size of training set. Also, the training set was over-sampled by adding random Gaussian noise to the latent variables. In short, the performance of the proposed model is here compared with the following four different approaches: 1) latent variables are extracted from the AE, without generating augmented data, and OCC algorithms are applied to detect anomalies; 2) latent variables are augmented by SMOTE and then OCC algorithms are used; 3) similar to the previous approach except ADASYN is used for augmenting the latent variables instead of SMOTE; 4) by adding random Gaussian noise the latent variables are augmented. In the proposed approach, latent AE variables are extracted from several late epochs in the training process, thus creating an augmented data set.

\par The fourth approach, i.e., augmenting data by adding artificial noise, was carried out to test whether the proposed approach was acting as a simple regularizer similar to input noise injection, or whether it represented a more distinctive contribution. The optimistic hypothesis was that by simply adding artificial noise it is not possible to achieve the same level of meaningful augmented data that can be obtained by the proposed approach. In other words, the assumption was that this simple transformation (i.e. adding noise) does not provide as much meaningful information as using latent representations from different epochs.

\par One of the most common evaluation metrics used in this domain is ROC AUC; however, authors in \cite{Provost:1997} claim that when the data set is highly imbalanced, ROC AUC is not a suitable metric and a good alternative would be the area under the precision-recall curve (PR AUC), especially when dealing with high dimensional data in which the positive class, i.e, anomalies, is more important than the negative class, i.e., normal points. Nonetheless, no single evaluation metric dominates others; therefore, both ROC AUC and PR AUC were used for evaluation. 

\par In Table \ref{tab:aucs_UNSW}, the performance of all four different models on the UNSW-NB15 data set is listed in terms of PR AUC and ROC AUC. The backgrounds of the cells with the highest values are highlighted in grey. In the UNSW-NB15 data set, there are 9 different types of anomalies and the experiment was carried out on each one individually. On each data set, the experiment was repeated 10 times. To eliminate the effect of extreme values \cite{Su2017}, the largest and smallest values were removed and the average performance was tabulated. As listed in Table \ref{tab:aucs_UNSW}, the proposed data augmentation method dominated both in terms of PR AUC and ROC AUC in almost every type of anomaly. By looking at the performance of models when applied to \textit{Exploits} and \textit{Reconnaissance}, in terms of ROC AUC, it can be seen that the proposed model came second after the approach in which artificial noise was used for data augmentation; however, by a small margin, i.e., 0.044 and 0.003 respectively. It is worth mentioning that all approaches performed competitively well on the \textit{Worms} data set while the proposed approach showed the second best result in terms of PR AUC.

\par As depicted in the Table \ref{tab:aucs_nlkdd_ctu13}, the proposed approach outperformed other approaches on NSL-KDD and CTU13 data sets. Also, it is noticeable that almost all the approaches performed equally well on the NSL-KDD (U2R) data set, especially in terms of PR AUC. It can be deduced that the extracted latent variables from the AE already make a good training set for OCC algorithms and it is fairly easy to compute the class boundary. Therefore, one can argue that data augmentation on similar scenarios is needless. 

\par Augmenting the training set using the AE completely dominated other approaches when applied on the data sets with a single-type anomaly, i.e., PenDigits, Spambase, InternetAds and Arrhythmia; except when applied to \textit{Shuttle} in which adding artificial noise with use of KDE showed a higher ROC AUC value. While LOF outperformed Isolation Forest and KDE when working with Arrhythmia, KDE showed superior results when applied to PenDigits and Spambase. Also, it is worth mentioning that even though the proposed method dominated when carried out on InternetAds, the performance shows a considerable margin when compared to other data sets. The assumption is that this is due to the high sparsity of this data set and the fact that it has the highest number of features, i.e., 1558, compared to other data sets. Thus, it requires a different AE to obtain useful latent variables without losing important information. However, in order to make sure the performance stays comparable to other data sets, it is important to maintain impartiality in every model. Therefore, the same approach for designing the AE architecture that was used for other data sets was followed on InternetAds. 

\begin{figure}
\includegraphics[width=\columnwidth]{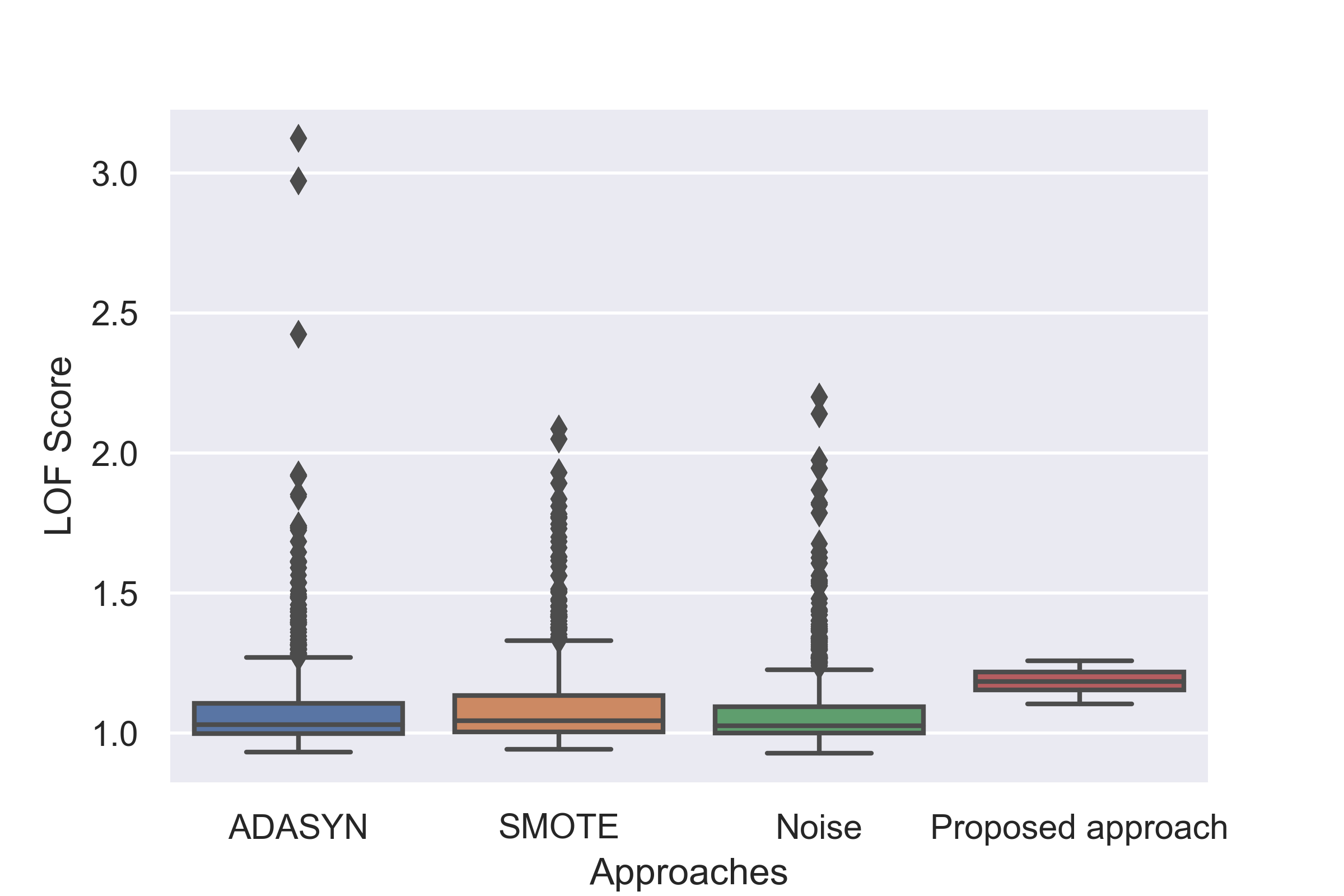}	
\caption{Boxplots of computed LOF scores for PenDigits data set after data augmentation with different approaches}
\label{fig:boxplot}
\end{figure}

\par In Fig. \ref{fig:boxplot}, a simple statistical analysis is carried out using Boxplots to see the effect of different data augmentation models on the LOF scores obtained from the PenDigits data set. Boxplots provide a good graphical representation, and as depicted in Fig. \ref{fig:boxplot}, it can be deduced that the other three approaches cause LOF to compute a wider range of LOF scores with some outliers while the proposed approach computed LOF scores that are within a small range. By looking at Fig. \ref{fig:boxplot} and also the performance of data augmentation with the proposed model and using LOF for detecting anomalies in Table \ref{tab:aucs_single_type}, it can be reasoned that the proposed model can augment the training set in a way that leads to finding a better class boundary for LOF. To verify whether the LOF scores are significantly different or not, a Wilcoxon signed-rank test \cite{kerby2014simple} was performed between the proposed method and the adding noise approach. By feeding LOF scores to the Wilcoxon test, it was observed that the results are significantly different at $p<0.05$. It is worth mentioning that carrying out repetitive Wilcoxon tests on multiple models is not recommended because it increases the chance of rejecting a certain proportion of the null hypotheses merely based on random chance \cite{demvsar2006statistical}. Therefore, the test was not carried out again with other approaches.

\section{Conclusion}
\label{section:conclusion}
\par This paper studies the usefulness of using AutoEncoders (AEs) for augmenting the training set in the feature-space improving the performance of One-Class Classifier (OCC) algorithms in anomaly detection problems. When the training set is not large enough, OCCs perform poorly as finding a good class boundary becomes difficult. Once the AE is almost well-trained, it is possible to start deriving latent variables in each epoch and feed this augmented training set to the OCC algorithm. 

\par By carrying out the proposed method on several well-known data sets in this domain, it was demonstrated that the proposed approach is capable of outperforming other data augmentation methods such as SMOTE and ADASYN. In terms of OCC methods, three state-of-the-art algorithms were employed to detect anomalies. According to the results, the proposed approach is capable of producing a more robust performance compared to other approaches.

\par For future work, it would be useful to investigate an adaptive approach for estimating the parameter $\nu$, which indicates the portion of epochs to use for augmentation. Also, it will be interesting to investigate the effectiveness of employing an ensemble of OCC algorithms with the proposed approach.

\bibliographystyle{ieeetr}
\bibliography{Oversampling_with_AE.bib}
\end{document}